\documentclass[runningheads]{llncs}
\usepackage{graphicx}

\usepackage{tikz}
\usepackage{comment}
\usepackage{color}

\usepackage[accsupp]{axessibility}  

\usepackage{array}
\usepackage[caption=false,font=normalsize,labelfont=sf,textfont=sf]{subfig}
\usepackage{textcomp}
\usepackage{stfloats}
\usepackage{url}
\usepackage{verbatim}
\usepackage{cite}
\usepackage{mathrsfs}
\usepackage{booktabs}
\usepackage{array}
\usepackage{colortbl}
\usepackage{wrapfig}
\usepackage{multirow}
\usepackage{multicol}
\usepackage[mathscr]{euscript}
\usepackage{ulem}
\usepackage{diagbox}
\usepackage[ruled,linesnumbered]{algorithm2e}

\usepackage[misc]{ifsym} 
\usepackage{bbding}

\makeatletter

\newcommand{\Rmnum}[1]{\expandafter\@slowromancap\romannumeral #1@}
\makeatother

\begin{document}

\pagestyle{headings}
\mainmatter

\title{1st Place Solution for ECCV 2022 OOD-CV Challenge Image Classification Track} 


\titlerunning{ECCV 2022 OOD-CV Challenge Image Classification Track}
\author{Yilu Guo\inst{1} 
\and
Xingyue Shi\inst{2,\footnotemark[1]}
\and
Weijie Chen \inst{1,3,\footnotemark[4]}
\and 
Shicai Yang \inst{1}
\and 
Di Xie \inst{1}
\and \\
Shiliang Pu \inst{1}
\and 
Yueting Zhuang \inst{3}
}

\authorrunning{Y. Guo et al.}
%
\institute{Hikvision Research Institute, Hangzhou China
\and
Peking University Shenzhen Graduate School, Shenzhen, China
\and
Zhejiang University, Hangzhou, China
\\
\email{\{guoyilu5, chenweijie5, yangshicai, xiedi, pushiliang.hri\}@hikvision.com, shixy@stu.pku.edu.cn, \{chenweijie, yzhuang\}@zju.edu.cn}
}

\maketitle
\renewcommand{\thefootnote}{\fnsymbol{footnote}}
\footnotetext[1]{Internship in Hikvision Research Institute}
\footnotetext[4]{Corresponding author}
\footnotetext[5]{https://www.ood-cv.org/}

\begin{abstract}
OOD-CV challenge\footnotemark[5] is an out-of-distribution generalization task. In this challenge, our core solution can be summarized as that \textbf{Noisy Label Learning Is A Strong Test-Time Domain Adaptation Optimizer}. Briefly speaking, our main pipeline can be divided into two stages, a pre-training stage for domain generalization and a test-time training stage for domain adaptation. We only exploit labeled source data in the pre-training stage and only exploit unlabeled target data in the test-time training stage. In the pre-training stage, we propose a simple yet effective \textbf{Mask-Level Copy-Paste} data augmentation strategy to enhance out-of-distribution generalization ability so as to resist shape, pose, context, texture, occlusion, and weather domain shifts in this challenge. In the test-time training stage, we use the pre-trained model to assign noisy label for the unlabeled target data, and propose a \textbf{Label-Periodically-Updated DivideMix} method for noisy label learning. After integrating Test-Time Augmentation and Model Ensemble strategies, our solution ranks the first place on the Image Classification Leaderboard of the OOD-CV Challenge. Code will be released in \url{https://github.com/hikvision-research/OOD-CV}.
\end{abstract}

\section{Introduction} 
\begin{figure}[!t]
    \centering
    \includegraphics[width=\linewidth]{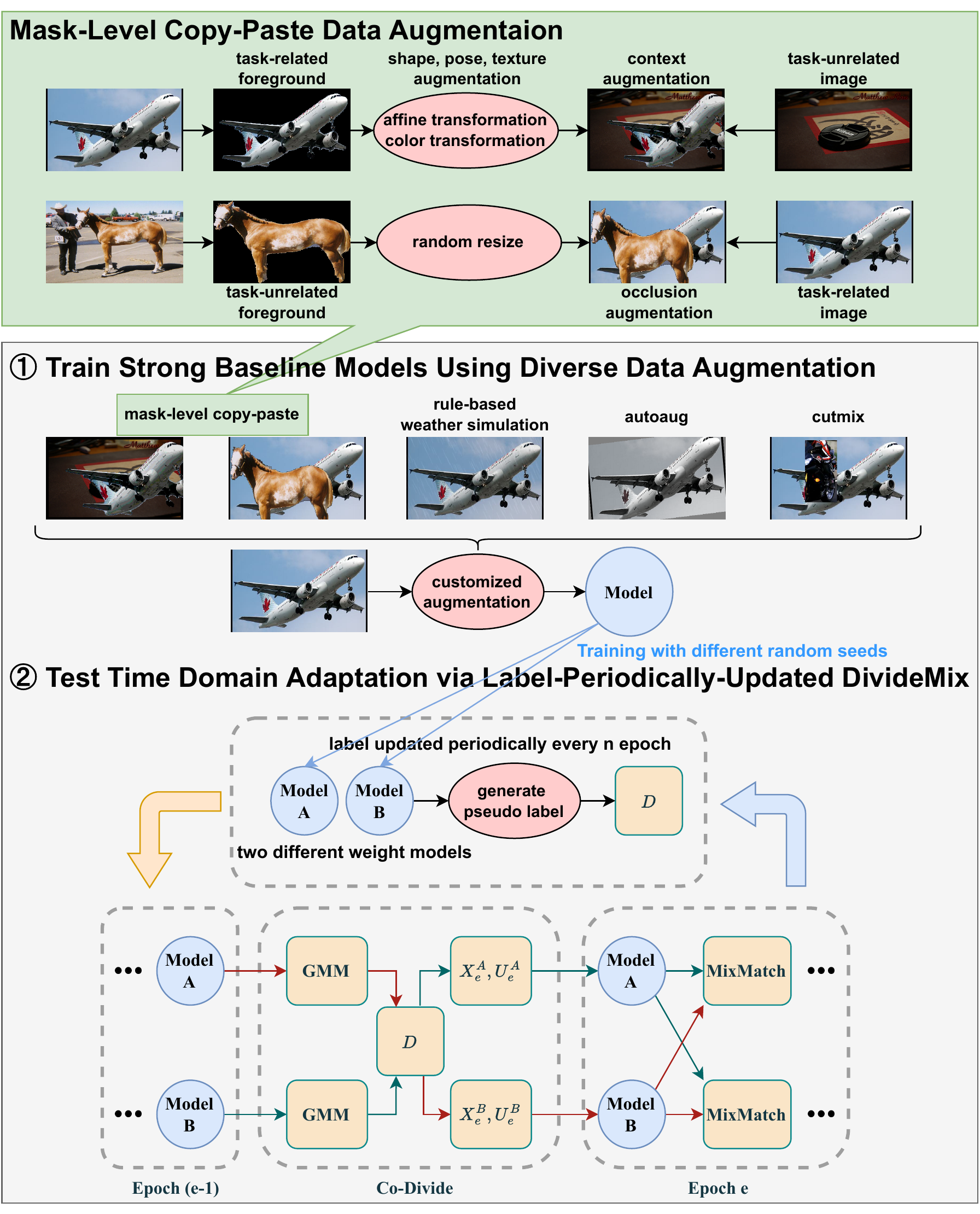}
    \vskip -0.1in
    \caption{The pipeline of our proposed framework, which is composed of a model pre-training stage by merely exploiting the labeled source data and a test-time domain adaptation stage by merely exploiting the unlabeled target data. Note that Mask-Level Copy-Paste is developed upon a weakly-supervised semantic segmentation method using only image-level label.}
    \label{dividemix}
\end{figure}

Following the existing works, such as SSNLL \cite{chen2021self} for Image Classification and SFOD \cite{li2021free} for Object Detection, we put forth the idea that \textbf{Noisy Label Learning Is A Strong Test-Time Domain Adaptation Optimizer}. 1) Before Test-Time Domain Adaptation, it is an essential prerequisite to pre-train a strong baseline model that generalizes well to the out-of-distribution data. An instinctual method is to stack diverse strong data augmentation strategies on the source data in order to resist diverse domain shifts. To this end, besides the conventional data augmentation, we develop a novel \textbf{Mask-Level Copy-Paste} data augmentation method. Specifically, given the image-level label, we adopt a state-of-the-art weakly-supervised semantic segmentation method MCTformer \cite{xu2022multi} to segment the foreground objects on the ImageNet-1K and ROBIN training datasets. In this way, we can apply affine transformation and color jittering to augment the foreground objects (against shape, pose and texture domain shifts), and then paste the task-related foreground objects onto task-unrelated image (against context domain shift) and paste the task-unrelated foreground objects onto task-related image (against occlusion domain shift). 2) After model pre-training, pseudo labels of target data inferred by the source pre-trained model can be viewed as the noisy labels. Under this consideration, existing noisy label learning methods, such as DivideMix \cite{li2020dividemix}, can be naturally used for test-time domain adaptation. In this challenge, we propose a \textbf{Label-Periodically-Updated DivideMix}, which can correct the noisy labels in time while avoiding overfitting the noisy labels. After integrating Test-Time Augmentation and Model Ensemble with diverse hyper-parameters, our solution ranks the first place on the Image Classification Leaderboard of the OOD-CV Challenge \url{https://codalab.lisn.upsaclay.fr/competitions/6781#results}.

Our main contribution can be summarized as follows:
\begin{itemize}
\item \textbf{Pretraining Stage: Mask-Level Copy-Paste.} We propose a novel Mask-Level Copy-Paste data augmentation strategy to resist diverse domain shifts.
\item \textbf{Adaptation Stage: Label-Periodically-Updated DivideMix.} We consider test-time training as a noisy label learning problem. We modify the DivideMix by updating noisy labels periodically and use strong-weak augmentation to get a significant performance improvement on the OOD data. 
\end{itemize}

\section{Method} 
\subsection{Mask-Level Copy-Paste.} We propose a novel Mask-Level Copy-Paste data augmentation strategy to resist diverse domain shifts. Specifically, we train a weakly-supervised semantic segmentation (WSSS) model~\cite{xu2022multi} by exploiting the image-level label on ImageNet-1K and ROBIN training datasets. We then segment the foreground objects by testing this WSSS model. According to whether the class label is related to the task in this challenge or not, the foreground objects can be divided into task-related and task-unrelated parts. Similarly, the images from ImageNet-1K and ROBIN training datasets can also be divided into task-related and task-unrelated parts. In this way, we can apply randomly affine transformation and color transformation to augment the foreground objects (against shape, pose and texture domain shifts), and then paste the task-related foreground objects onto task-unrelated image (against context domain shift) and paste the task-unrelated foreground objects onto task-related image (against occlusion domain shift). By stacking with other data augmentation strategies, including AutoAug~\cite{autoaug}, CutMix~\cite{cutmix} and rule-based weather simulation methods~\cite{imagenet-c} (against weather domain shift), we can get a pre-trained model with a strong domain generalization ability.

\subsection{Label-Periodically-Updated DivideMix.} We consider test time domain adaptation as a noisy label learning problem. And we regard the strong pre-trained model as a noisy annotator on the unlabeled test set. After obtaining noisy labels, we modify the DivideMix \cite{li2020dividemix} into a label-periodically-updated one, which can correct the noisy labels in time while avoiding overfitting the noisy labels. 
Furthermore, different from the original DivideMix, we employ the popular strong-weak augmentation in the MixMatch component, which uses weak augmentation for pseudo labeling and strong augmentation for model optimization.

\section{Technique Details}
\subsection{Dependencies}
\begin{itemize}
    \item \textbf{Transformer Backbone}: Swinv2 \cite{swinv2}
    \item \textbf{Dataset}: ROBIN \cite{zhao21robin}, ImageNet-1K \cite{Russakovsky2015ImageNetLS}
    \item \textbf{Data Augmentation}: AutoAug \cite{autoaug}, CutMix \cite{cutmix}, Rule-Based Weather Simulation \cite{imagenet-c}
    \item \textbf{Weakly-Supervised Semantic Segmentation}: MCTformer \cite{xu2022multi}
    \item \textbf{Noisy Label Learning Method}: DivideMix \cite{li2020dividemix}
\end{itemize}

\subsection{Training Description}
\subsubsection{Baseline model.}
Comparing different model architectures, we select Swinv2-B \cite{swinv2} pretrained on the ImageNet-1K dataset as the baseline
to develop our method. Swinv2 is a hierarchical vision Transformer whose representation is determined by a shifted windowing scheme, which introduces locality to the model and improves computation efficiency.

\subsubsection{Data augmentation.}
1) We design a \textbf{Mask-Level Copy-Paste} data augmentation strategy against different nuisances based on weakly-supervised semantic segmentation. An image-level supervised semantic segmentation model is trained on the ROBIN training dataset via MCTformer \cite{xu2022multi}. After getting the segmented masks, we can apply randomly affine transformation (object distortion, rotation, horizontal and vertical flipping) and color transformation to augment the foreground objects (against shape, pose and texture domain shifts), and then paste the task-related foreground objects onto task-unrelated image (against context domain shift). In addition, we train another image-level supervised semantic segmentation model on the images from task-unrelated classes in the ImageNet-1K dataset. Then we randomly paste the task-unrelated foreground objects onto task-related image to imitate object occlusion (against occlusion domain shift). 2) Rule-based weather simulation. We follow ImageNet-C \cite{imagenet-c} to implement four image transformations simulating four common weather conditions, including rain, snow, fog and sunshine (against weather domain shift). The detailed implementation is based on three Python libraries, including \texttt{cv2}, \texttt{skimage} and \texttt{scipy.ndimage}. 3) We also adopt two common data augmentation strategies like AutoAug \cite{autoaug} and CutMix \cite{cutmix}. By stacking multiple data augmentations, we can obtain a {\tt Strong Baseline} model.

\subsubsection{Implementation details.}
We use the cross-entropy loss with label smoothing as the loss function, and the smoothing factor is set to $ 0.1 $. Swinv2-B model is trained with 8 GPUs and 32 samples per GPU. We use the SGD optimizer with momentum 0.9. The learning rate is adjusted according to the cosine decaying policy [8] and the initial learning rate is set to 0.01. The warm-up [8] strategy is applied over the first 3 epochs, gradually increasing the learning rate linearly from 1e-6 to the initial value of the cosine schedule. The weight decay is set to 2e-5. The model is trained for 100 epochs and is stopped when the accuracy on the validation set no longer increases.

\subsection{Testing Description}
\subsubsection{Test-time domain adaptation.}
The classifier trained on the training set suffers great performance degradation on the out-of-distribution test set. Regarding the trained classifier as a noisy annotator on the unlabeled test set, the test-time training procedure can be considered as an instance of noisy-label learning. To this end, DivideMix \cite{li2020dividemix} is modified to conduct test-time noisy-label learning as shown in Fig.\ref{dividemix}. Firstly, the classifier trained with only training set assigns noisy pseudo-labels to the unlabeled test set, denoted as the test-time training set. Then, two models (A and B) are trained by different random seeds or hyper-parameters on the training set for co-teaching. Every epoch, a model divides the noisy training set into a labeled subset (mostly clean) $ X $ and an unlabeled subset (mostly noisy) $ U $ based on the assumption that clean samples have lower losses. DivideMix performs label co-refinement on $ X $ and label co-guessing on $ U $ to enable the two models to teach each other. After obtaining $ \hat{X} $ (and $ \hat{U} $) which contain multiple augmented views of labeled (and unlabeled) samples and their refined (and guessed) labels, DivideMix follows MixMatch \cite{mixmatch} to mix-up $ \hat{X} $ and $ \hat{U} $, and calculate the cross-entropy loss and the MSE (mean square error) loss on them respectively. Specifically, we use weak augmented views for getting refined (guessed) labels and use strong augmented views (AutoAug) for training. During DivideMix, the performance might degenerate after a number of epochs. We update noisy pseudo-labels every 3 epochs, which can correct the noisy labels in time while avoiding overfitting the noisy labels. Also, we rebuild the source code of DivideMix project from single-GPU to multi-GPU to speed up training.

\subsubsection{Implementation details.}
During test time training, only AutoAug is utilized for strong data augmentation. The model are trained with 8 GPUs and 6 samples per GPU. We use the SGD optimizer with momentum 0.9. The learning rate is set to 0.02 and the weight decay is set to 5e-4. To get the labeled subset $ X $ from the noisy test-time training set, we set the clean probability threshold ({\tt p}) to 0.8, which is higher than the default value 0.5 in the original DivideMix \cite{li2020dividemix}, so as to achieve more reliable pseudo-labeled samples.

\subsubsection{Testing phase.}
During testing, the TenCrop inference are utilized which crops the given image into four corners and the central crop plus the horizontal flipped version of these crops.

\section{Experimental Results}
\subsection{Ablation Study}
\begin{table}[h!]
\centering
\resizebox{.98\textwidth}{!}{
\begin{tabular}{lcccccccc}
\hline
\multirow{2}{*}{Methods}& \multirow{2}{*}{IID} & \multicolumn{7}{c}{OOD} \\ \cline{3-9}
&&Shape & Pose & Context & Texture & Occlusion & Weather & Avg.  \\ \hline
Swinv2-B & 90.90 & 85.97 & 90.56 & 87.67 & 95.25 & 83.63 & 92.09 & 89.19\\ \hline
Swinv2-B + AutoAug + CutMix & 90.94 & 86.99& 90.46& 88.73& 95.77& 86.32& 92.38 & 90.11\\ \hline
Strong Baseline & 90.80& 87.09 & 90.56 & 89.26 &95.25 & 91.74 & 93.34 & 91.21\\ \hline
Strong Baseline + TTT & 87.10 & 88.99 & 93.77 & 96.30 & 96.81 & 93.84 & 95.18 & 94.15 \\ \hline
Strong Baseline + TTT + TenCrop & 87.10 & 89.40 & 94.23 & 97.18 & 97.35 & 94.96 & 95.55 & 94.78\\ \hline
\end{tabular}
}
\vskip 0.05in
\caption{Some ablation results on phase 1 of the challenge. ``TTT'' is short for Test-Time Training. ``Avg.'' is the average accuracy of the six domains.}
\label{table:phase1}
\end{table}

Table \ref{table:phase1}  shows the ablation results on phase 1 of the challenge. The techniques can improve the OOD accuracy from 89.19\% to 94.78\%. It is much higher than the performance (92.65\%) on the leaderboard of phase 1. Since there are more OOD data than IID data in the test set, the IID performance degenerates after test time training.

\subsection{Ensembles And Fusion Strategies}

\begin{table}[h!]
\centering
\resizebox{.98\textwidth}{!}{
\label{table:phase2}
\begin{tabular}{lcccccccc}
\hline
\multirow{2}{*}{Settings}& \multirow{2}{*}{IID} & \multicolumn{7}{c}{OOD} \\ \cline{3-9}
&&Shape & Pose & Context & Texture & Occlusion & Weather & Avg.  \\ \hline
p=0.85 & 85.40 & 84.09 & 84.81 & 87.39 & 76.93 & 95.02 & 85.74 & 85.74 \\ \hline
p=0.85 w/ a different seed & 90.22 & 84.77 & 82.93 & 86.08 & 73.96 & 97.32 & 86.60 & 85.28 \\ \hline
p=0.8 & 89.23 & 84.09 & 85.62 & 87.46 & 75.21 & 97.73 & 87.60 & 86.29 \\ \hline
p=0.8 w/ a different epoch & 90.88 & 83.27 & 85.89 & 87.46 & 77.10 & 97.70 & 86.40 & 86.30 \\ \hline
Model ensemble & 90.68 & 85.11 & 85.48 & 88.28 & 77.50 & 97.99 & 87.80 &  87.03\\ \hline
\end{tabular}
}
\vskip 0.05in
\caption{Final results on phase 2 of the challenge. The 2nd-row model is set as the anchor model. The last-row (6th-row) result is the ensemble from the models in 2nd-5th rows via an entropy-guided model adaptive ensemble strategy. Note that {\tt p} is a hyper-parameter in DivideMix to split noisy data into a labeled set and an unlabeled set for MixMatch training.}
\end{table}

We ensemble 4 models trained by different hyper-parameters via an entropy-guided model adaptive ensemble strategy. Specifically, we choose a model as an anchor model. For each image, we trust the result if the entropy of the prediction from the anchor model is small enough (less than 2/3 mean entropy of the test set). Otherwise, we use the results from 4 models for ensemble. In this way, the inference time can be reduced. The 4 models are trained by different random seeds or hyper-parameters, different {\tt p} and different training epochs in DivideMix. After ensemble, the OOD accuracy is improved from 86.30\% to 87.03\%.

\subsection{Model Complexity}

\begin{table}[h!]
\centering
\begin{tabular}{cccc}
\hline
& complexity & Parameters & Train Time \\ \hline
Single model & 22.01GFLOPs & 86.9M & 12hours * 8GPUs\\ \hline
TenCrop+Ensemble & 880.4GFLOPs & 347.6M & 48hours * 8GPUs\\ \hline
\end{tabular}
\vskip 0.05in
\caption{Method complexity analysis. Four models with different training hyper-parameters are used for model ensemble. Eight Nvidia Tesla V100 GPUs are used for testing the Train Time.}
\label{table:descrip}
\end{table}

Please refer to Table \ref{table:descrip} for model complexity analysis.

\section{Conclusion}
The OOD-CV challenge deepens our understanding of model robustness under domain shifts. We thank the organizers for their great work in the challenge.

\bibliographystyle{splncs04}
\bibliography{egbib}

\begin{thebibliography}{10}
\providecommand{\url}[1]{\texttt{#1}}
\providecommand{\urlprefix}{URL }
\providecommand{\doi}[1]{https://doi.org/#1}

\bibitem{mixmatch}
Berthelot, D., Carlini, N., Goodfellow, I., Papernot, N., Oliver, A., Raffel,
  C.A.: Mixmatch: A holistic approach to semi-supervised learning. In: Wallach,
  H., Larochelle, H., Beygelzimer, A., d\textquotesingle Alch\'{e}-Buc, F.,
  Fox, E., Garnett, R. (eds.) Advances in Neural Information Processing
  Systems. vol.~32. Curran Associates, Inc. (2019),
  \url{https://proceedings.neurips.cc/paper/2019/file/1cd138d0499a68f4bb72bee04bbec2d7-Paper.pdf}

\bibitem{chen2021self}
Chen, W., Lin, L., Yang, S., Xie, D., Pu, S., Zhuang, Y.: Self-supervised noisy
  label learning for source-free unsupervised domain adaptation. In: IROS
  (2022)

\bibitem{autoaug}
Cubuk, E.D., Zoph, B., Man{\'{e}}, D., Vasudevan, V., Le, Q.V.: Autoaugment:
  Learning augmentation policies from data. CoRR  \textbf{abs/1805.09501}
  (2018), \url{http://arxiv.org/abs/1805.09501}

\bibitem{imagenet-c}
Hendrycks, D., Dietterich, T.: Benchmarking neural network robustness to common
  corruptions and perturbations. Proceedings of the International Conference on
  Learning Representations  (2019)

\bibitem{li2020dividemix}
Li, J., Socher, R., Hoi, S.C.: Dividemix: Learning with noisy labels as
  semi-supervised learning. In: International Conference on Learning
  Representations (2020)

\bibitem{li2021free}
Li, X., Chen, W., Xie, D., Yang, S., Yuan, P., Pu, S., Zhuang, Y.: A free lunch
  for unsupervised domain adaptive object detection without source data. In:
  Proceedings of the AAAI Conference on Artificial Intelligence. pp. 8474--8481
  (2021)

\bibitem{swinv2}
Liu, Z., Hu, H., Lin, Y., Yao, Z., Xie, Z., Wei, Y., Ning, J., Cao, Y., Zhang,
  Z., Dong, L., Wei, F., Guo, B.: Swin transformer v2: Scaling up capacity and
  resolution. In: 2022 IEEE/CVF Conference on Computer Vision and Pattern
  Recognition (CVPR). pp. 11999--12009 (2022).
  \doi{10.1109/CVPR52688.2022.01170}

\bibitem{Russakovsky2015ImageNetLS}
Russakovsky, O., Deng, J., Su, H., Krause, J., Satheesh, S., Ma, S., Huang, Z.,
  Karpathy, A., Khosla, A., Bernstein, M.S., Berg, A.C., Fei-Fei, L.: Imagenet
  large scale visual recognition challenge. International Journal of Computer
  Vision  \textbf{115},  211--252 (2015)

\bibitem{xu2022multi}
Xu, L., Ouyang, W., Bennamoun, M., Boussaid, F., Xu, D.: Multi-class token
  transformer for weakly supervised semantic segmentation. In: Proceedings of
  the IEEE/CVF Conference on Computer Vision and Pattern Recognition. pp.
  4310--4319 (2022)

\bibitem{cutmix}
Yun, S., Han, D., Chun, S., Oh, S.J., Yoo, Y., Choe, J.: Cutmix: Regularization
  strategy to train strong classifiers with localizable features. In: 2019
  IEEE/CVF International Conference on Computer Vision (ICCV). pp. 6022--6031
  (2019). \doi{10.1109/ICCV.2019.00612}

\bibitem{zhao21robin}
Zhao, B., Yu, S., Ma, W., Yu, M., Mei, S., Wang, A., He, J., Yuille, A.,
  Kortylewski, A.: Robin: A benchmark for robustness to individual nuisances in
  real-world out-of-distribution shifts. arXiv preprint arXiv:2111.14341
  (2021)

\end{thebibliography}
\end{document}